\documentclass[letterpaper, 10 pt, conference]{ieeeconf}  

\IEEEoverridecommandlockouts                              

\usepackage{graphicx}
\usepackage{subfigure}
\usepackage{cite}
\usepackage{amsmath,amssymb,amsfonts}
\usepackage{tabularx,ragged2e,booktabs}
\usepackage{multirow}
\usepackage{tikz}
\usepackage{textcomp}
\usepackage{hyperref}
\usepackage{lipsum}

\newcommand\copyrighttext{%
  \footnotesize \textcopyright 2021 IEEE. Personal use of this material is permitted. Permission from IEEE must be obtained for all other uses, in any current or future media, including reprinting/republishing this material for advertising or promotional purposes, creating new collective works, for resale or redistribution to servers or lists, or reuse of any copyrighted component of this work in other works.}
\newcommand\copyrightnotic{%
\begin{tikzpicture}[remember picture,overlay]
\node[anchor=south,yshift=10pt] at (current page.south) {\fbox{\parbox{\dimexpr\textwidth-\fboxsep-\fboxrule\relax}{\copyrighttext}}};
\end{tikzpicture}%
}

\newcommand{\norm}[1]{\left\lVert #1 \right\rVert}

\title{\LARGE \bf
Adversarial Attacks on Camera-LiDAR Models for 3D Car Detection 
}

\author{Mazen Abdelfattah$^{1}$, Kaiwen Yuan$^{1}$, Z. Jane Wang$^{1}$, and Rabab Ward$^{1}$

\thanks{$^{1}$Mazen Abdelfattah, Kaiwen Yuan, Z. Jane Wang, and Rabab Ward are with the Department of Electrical and Computer Engineering, University of British Columbia, Vancouver, BC, Canada (email: \{mazen,  kaiwen, zjanew, rababw\}@ece.ubc.ca;). \emph{Corresponding author: Mazen Abdelfattah.}}%
}

\begin{document}

\maketitle
\copyrightnotic
\thispagestyle{empty}
\pagestyle{empty}

\begin{abstract}
Most autonomous vehicles (AVs) rely on LiDAR and RGB camera sensors for perception. Using these point cloud and image data, perception models based on deep neural nets (DNNs) have achieved state-of-the-art performance in 3D detection. The vulnerability of DNNs to adversarial attacks has been heavily investigated in the RGB image domain and more recently in the point cloud domain, but rarely in both domains simultaneously. Multi-modal perception systems used in AVs can be divided into two broad types: cascaded models which use each modality independently, and fusion models which learn from different modalities simultaneously. We propose a universal and physically realizable adversarial attack for each  type, and study and contrast their respective vulnerabilities to attacks. We place a single adversarial object with specific shape and texture on top of a car with the objective of making this car evade detection. Evaluating on the popular KITTI benchmark, our adversarial object made the host vehicle escape detection by each model type more than 50\% of the time. The dense RGB input contributed more to the success of the adversarial attacks on both cascaded and fusion models.
\end{abstract}


\vspace{-0.2cm}
\section{INTRODUCTION}
Autonomous vehicles (AVs) employ RGB cameras and LiDAR sensors to generate complimentary representations of the scene in the form of dense 2D RGB images and sparse 3D point clouds. Using these data, 3D car detection models based on deep neural networks (DNNs) have achieved state-of-the-art performance. Images and point clouds are utilized differently depending on the model type (cascaded, or fusion) as shown in Fig. \ref{pipe_frustum}, and \ref{pipe_fusion}. Cascaded models use each modality independently. Images are used by a 2D detection DNN to generate proposals of search spaces where a car may reside. LiDAR points in each proposed region are then extracted for 3D point-based detection \cite{frustum, roarnet}. On the other hand, fusion models \cite{epnet, mvx} use DNNs to extract and fuse image and point cloud features in parallel. Then, a combined representation is sent through a DNN for 3D detection. 

\begin{figure}[htbp]
\centering
  \subfigure{
      \includegraphics[width=0.38\linewidth]{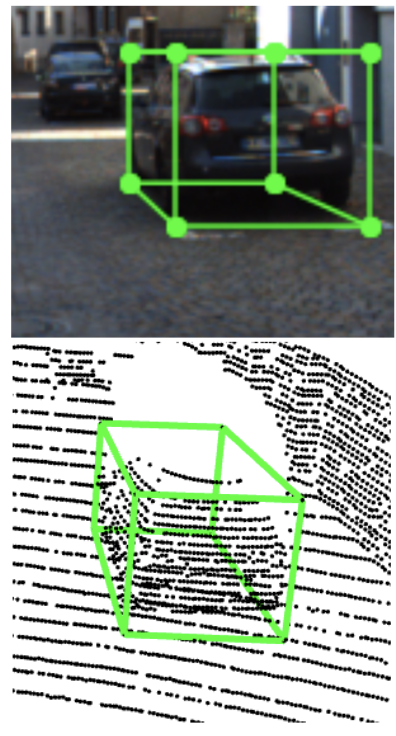}}
         \subfigure{
        \includegraphics[width=0.34\linewidth]{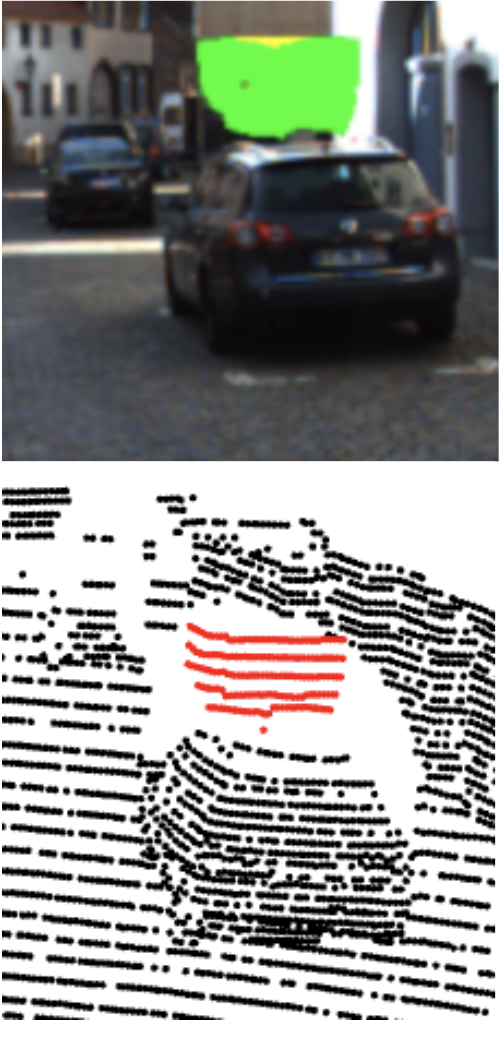}}
        \vspace{-0.2cm}
\caption{\hspace{-0.05cm}Left: A car is detected with accurate bounding boxes. Right: The same car is not detected after the adversarial attack. We place a 3D adversarial mesh with learned shape and texture on top of a car. The adversarial texture fools the 2D detection pipeline, and the red points are the rendered LiDAR points added to the point cloud to fool the 3D pipeline.}
\label{fig1}
\vspace{-0.6cm}
\end{figure}

The aforementioned models rely on DNNs which are known to be vulnerable to adversarial attacks that slightly alter the input but greatly affect the output. These attacks were initially observed and studied in the RGB image domain \cite{adversarial, patch, sharif2016accessorize}. Realistic adversarial attacks on 3D detection models that take only point clouds as input were recently investigated in \cite{mich2019, uber}. They placed a perturbed mesh in the 3D scene and rendered it differentiably by simulating a LiDAR. However, this prior work only considered one modality, while most AVs rely on both modalities (images and point clouds) for perception. As safety is critical for AVs, this paper investigates realistic adversarial attacks on detection models that take both images and point clouds as input, specifically we consider the common cascaded and fusion architectures.

\begin{figure*}[t]
 \center
 \vspace{0.2cm}
  \includegraphics[scale=0.5]{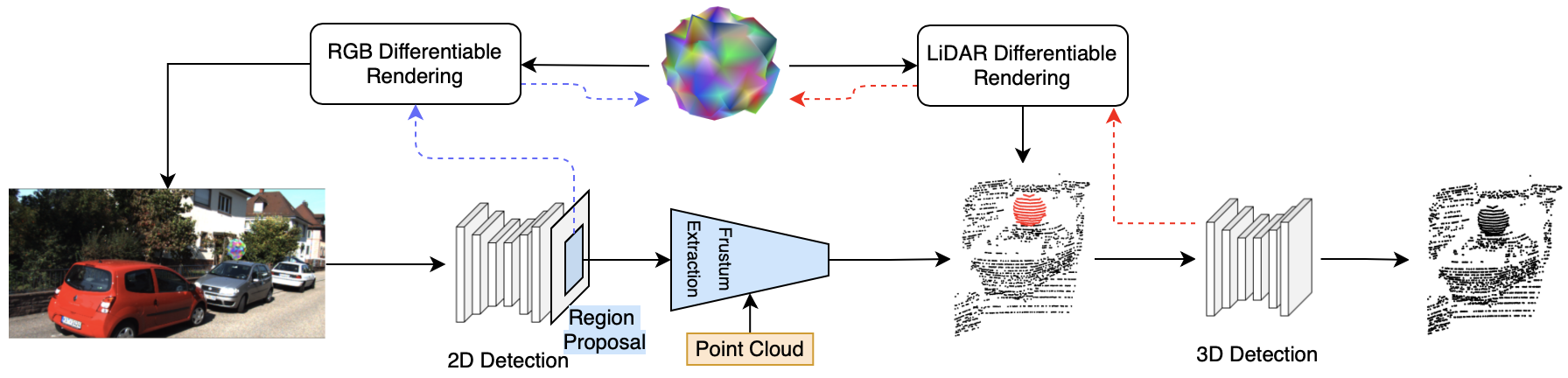}
  \vspace{-0.3cm}.
  \caption{The proposed attack pipeline on the cascaded F-PN model \cite{frustum}. Gradient flow is represented by dashed lines. Purple is for texture changing gradients, and red is for shape changing gradients. The car with the adversarial object goes undetected.}
  \label{pipe_frustum}
  \vspace{-0.5cm}
\end{figure*}

Recently, \cite{uber2} used differntiable rendering to make a multi-sensor adversarial attack on the fusion-based bird's eye view (BEV) detector ''Multi-task multi-sensor fusion" (MMF)\cite{mmf}. While \cite{uber2} has considered fusion models, we study adversarial attacks on both cascaded and fusion camera-LiDAR models in the context of AVs. In contrast to \cite{uber2} which attacks a model that is not publicly available, the two models we consider are both open-source and built upon commonly used backbones -as will be explained in section \ref{sec:methods}- which can make our results more generalizable.

We propose a universal, and realistic adversarial attack on cascaded and fusion 3D car detection models. The malicious adversarial object is placed on top of a car to avoid occlusion (see Fig. \ref{fig1}). This object is then \emph{digitally} rendered to point cloud and the corresponding RGB image by differentiable renderers, thus maintaining consistent and realistic characteristics across inputs. The shape and texture of the adversarial object are trainable parameters that are perturbed adversarially. This adversarial object is trained on the entire KITTI \cite{kitti} dataset which makes our attack universal, i.e. this single object is capable of reducing the accuracy of a perception system in different 3D scenes. For the cascaded type, we chose Frustum-PointNet (F-PN) \cite{frustum} which is a pioneering work in cascaded models and many more works were based on its architecture. As for the fusion type, we chose EPNet \cite{epnet} which is a powerful multi-modal 3D detector. EPNet works directly on LiDAR points (not their projection to BEV) to avoid information loss due to projection. EPNet \cite{epnet} focuses on the 3D detection task and is more accurate than MMF \cite{mmf} in 3D and BEV detection. 

The proposed attack shows how this adversarial object can make its host vehicle evade detection. This is especially dangerous in the context of self-driving cars where car detection accuracy is consequential in autonomous decision-making. We not only report the attack success rate in hiding cars, but also the reduction in the average precision after an attack on all input vehicles, to aid in comparing our results to previous and future work on adversarial attacks and defenses. We find that both models were vulnerable to attacks that alter the input image. This can be due to the dense and brittle nature of RGB features. Moreover, attacks that target only the point cloud input are much more effective on the cascaded model than the fusion one. This is probably because fusion models directly incorporate image features which can lend robustness against LiDAR-only adversarial attacks.

\section{RELATED WORK}

Adversarial attacks were first discovered and studied in RGB image classification DNNs \cite{adversarial}. These attacks would change the pixel values of an image slightly and lead to great errors in the DNN output. However, such pixel-wise attacks were not realistic and would usually work on a single image. Therefore, constraints were established to ensure the physical feasibility of an attack \cite{patch, sharif2016accessorize}. Also, universal attacks were proposed where a single perturbation was effective across the input distribution \cite{moosavi2017universal}. 
Our attack is very different from the aforementioned work: We aim not to learn a patch or particular pixel values, but an adversarial texture on a 3D object that can attack a model when rendered to RGB images. Moreover, we target both the image and the point cloud input. Finally, some works proposed using a camouflage pattern on the entire vehicle to hide it from 2D detection \cite{zhang2018camou}. This approach is far more conspicuous than ours, and often simulates an entire scene. Instead we use real-life scenes and only simulate the small adversarial object making our attacks a little more stealthy and closer to reality.

Early work on adversarial attacks on point cloud models focused on perturbing point clouds by slightly moving, adding, or removing a few individual points \cite{yang2019adversarial, Xiang_2019_CVPR}. Due to LiDAR properties such perturbations were largely unrealizable physically. Towards a realistic point cloud attack on detection models in AVs, \cite{mich2019, uber} used an adversarial 3D mesh that is placed somewhere in the scene around an AV. Other methods would use LiDAR sensor spoofing to attack a model \cite{cao2019adversarial}. Such attacks however can be difficult to replicate considering how dynamic an AV's scene often is. These prior approaches still focus only on LiDARs, while an AV usually utilizes cameras as well.

In \cite{wang2020towards}, Wang et al. study adversarial attacks on multi-modal DNNs by shifting individual points in the 3D space and using a 2D RGB sticker. As stated earlier, simple point perturbations can be unrealistic due to LiDAR sensor properties, and this work ignores cross-modality consistency. Concurrently with our work, \cite{uber2} used a single 3D object to attack a fusion-based BEV detector. While \cite{uber2} focused on fusion models, we additionally consider cascaded models, which are more interpretable and less computationally expensive than fusion models \cite{fusionrev}. This makes cascaded models easier to train, test, and deploy in industry and thus more critical to examine, while fusion models are relatively less mature. In addition to reporting attack success rates like \cite{uber2}, we also report the overall reduction in the car detection accuracy of a multi-modal detector. This allows us to compare with the detector's initial accuracy as well as other attack or defense methods. Finally, their victim model and their attack are based on BEV detection which is different from our 3D detection task (e.g., BEV detection does not estimate 3D bounding boxes which are necessary for AVs). Also, BEV learning is done on projected point clouds which leads to information loss. This can make their attack results harder to extend to fusion multi-modal DNNs that reason on LiDAR points directly like \cite{epnet}.

\begin{figure*}[t]
 \center
 \vspace{0.2cm}
  \includegraphics[scale=0.43]{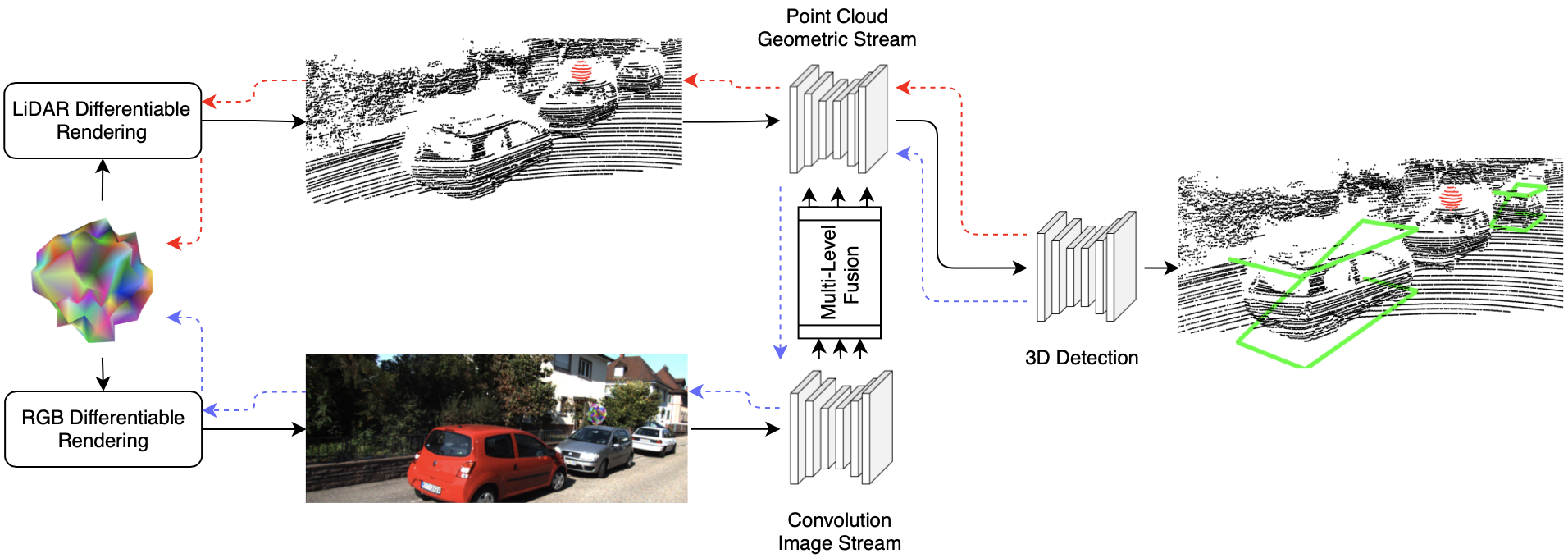}
  \vspace{-0.3cm}.
  \caption{The proposed attack pipeline on the fusion EPNet model \cite{epnet}. The gradient flow is represented by dashed lines. Purple is for texture changing gradients, and red is for shape changing gradients. The car with the adversarial object goes undetected.}
  \label{pipe_fusion}
  \vspace{-0.5cm}
\end{figure*}

\vspace{-0.3cm}
\section{Multi-Modal Adversarial Attack}
\vspace{-0.3cm}
\label{sec:methods}

In this work, we want to learn a single 3D adversarial object that is placed on top of a car in a 3D scene, to avoid occlusion, with the aim of dodging detection. This object is rendered to both point cloud and RGB image as if it was present in the original scene (see Fig.\hspace{4pt}\ref{pipe_frustum}, \ref{pipe_fusion}). We use this attack to study how this adversarial object affects the car detection accuracy of an AV's camera-LiDAR 3D detection model. We focus on car detection since it is one of the most safety-critical tasks of an AV's perception system. Below we elaborate on the adversarial object, differentiable rendering, the victim models, and the attack objective functions.

\subsection{Adversarial Object}

To craft a realistic adversarial object we choose a mesh as a graphical representation. A mesh can maintain realistic 3D physical geometry and texture and can also be differentiably rendered to RGB images and point cloud. 

To train the shape of the object, following previous work \cite{mich2019, uber}, we start with an initial mesh $S$ with $V$ vertices, where each initial vertex is defined as $v_i^0 \in \mathbb{R}^3 \text{, } i \in \{1, 2, ..., V\}$. We deform the shape of the mesh by displacing each vertex with a displacement vector $\hat{v}_i\in \mathbb{R}^3$, which is a learnable parameter to produce an adversarial mesh $S^{adv}$ with vertices  $v_i \in \mathbb{R}^3$, as in Eqn. \eqref{deform}. We use bounding box location and orientation to make a $4\times4$ transformation matrix $\mathbf{T}$ that puts the mesh on top of a host car and gives it the same orientation as the car's heading direction. 
\vspace{-0.15cm}
\begin{equation}
\vspace{-0.15cm}
    v_i = \mathbf{T} \cdot (v_i^0 + \hat{v}_i ) \label{deform}
\end{equation}

For the texture of the adversarial mesh, a learnable vector $c_i \in \mathbb{R}^3$ is assigned to each vertex to represent an RGB color.  To produce the texture of the mesh, each face is colored by the interpolation of the colors from its three vertices. The mesh is then rendered and added to the 2D RGB camera space with the given projection matrix.

To ensure the object is realistic we put constraints on the size and smoothness of the mesh geometry and interpolate between vertex colors to produce the adversarial texture. The result is a realistic smooth texture without abrupt changes in color (a typical observation in RGB adversarial attacks using pixel-wise perturbations or dense texture representations).

\vspace{-0.1cm}

\subsection{Differentiable Rendering}
To realistically render a mesh to point cloud, we need to find which LiDAR rays would intersect with the mesh if it was placed in the original scene. We simulate the LiDAR used in capturing the dataset's point cloud by producing rays at the same angular frequencies and incorporating a small amount of noise that is present in this specific LiDAR. We then calculate the intersection points between these rays and our mesh's faces in the 3D scene using the Möller–Trumbore intersection algorithm \cite{raycast}. Finally, for each ray with intersection points we take the nearest point, and add all the resulting points to the LiDAR point cloud scene.

 To train our adversarial texture, the rendering process needs to be differentiable. We therefore use the fast, differentiable rendering tools developed in \cite{ravi2020pytorch3d} to render our adversarial mesh from the 3D scene to RGB images. For image rendering, we use Phong shading and simulate sunlight by pointing a light source from above at a $45^\circ$ angle.

\subsection{Victim Models} 
Frustum-PointNet (F-PN) \cite{frustum} and EPNet \cite{epnet} are the representative cascaded and fusion victim models respectively. We chose F-PN \cite{frustum} for many reasons: It is a pioneering model in the area of multi-modal detection for AVs and other works \cite{roarnet} were developed based on its architecture and popular backbone DNN \cite{pointnet}. This means our attack could be a threat to other similar cascaded camera-LiDAR 3D detection models. Finally, F-PN showed competitive results on the KITTI benchmark.

As shown in Fig.\hspace{4pt}\ref{pipe_frustum}, F-PN deals with each input modality separately. First, it takes an RGB image through a 2D detection DNN, which proposes 2D bounding boxes. These bounding boxes are then projected to 3D space, thus producing frustum-shaped 3D search spaces that surround each object. Points within each frustum are extracted and sent through two PointNet-based  \cite{pointnet} DNNs for 3D instance segmentation, and 3D box estimation. F-PN directly outputs one 3D bounding box estimation from a given point cloud frustum, since the assumption is that there must be a single object in a frustum. This implicit bias can pose a challenge for adversarial attacks that seek to suppress detection by targeting only the point cloud pipeline. Also, the post-proposal point cloud is very sparse and small in size, therefore there is not much maneuvering space for our realistic adversarial attack to take place. Despite these challenges our attack was successful as discussed in section \ref{sec:experiments}.

For the image detection pipline in F-PN, we use YOLOv3 \cite{yolov3}. It is a standard fast 2D object detection DNN that outputs region proposals, classifications, and confidence scores. The F-PN and YOLOv3 models were pre-trained on the KITTI dataset. 

From the fusion architecture we attack the multi-modal 3D detection model EPNet \cite{epnet}. Unlike the cascaded model, this model takes both image and point cloud inputs simultaneously, and then learns and fuses their features in parallel. EPNet uses a novel fusion module that enhances point features with semantic image features at multiple scales down the network. Its image stream is built on standard 2D convolutional neural nets and its point cloud stream is based on modules developed in \cite{pointnet++}. EPNet uses the enhanced point features to generate many 3D bounding box proposals for each foreground point of a car and then uses a refinement network to filter proposals and refine bounding box dimensions. We note that there are some similarities between this network and the LiDAR only model PointRCNN \cite{pointrcnn}. For example, PointRCNN also generates proposals from each foreground point of a car and thus generates many proposals for each car. Due to that, a previous point cloud-only adversarial attack \cite{uber} found it very difficult to suppress all these proposals in their attack. While EPNet's usage of image features gave it a higher car detection accuracy than PointRCNN, it can be interesting to see how incorporating image features affects the robustness of a multi-modal model that is similar to a relatively robust LiDAR-only model.

\vspace{-0.1cm}
\subsection{Objective Functions}
In F-PN we need two separate objective functions since 2D and 3D detection are independent. For 2D detection, we minimize the sum of all objectness scores given to car detections using binary cross entropy: ${\mathcal{L}^{yolo} = -1 \cdot log(1-o_s)}$. For 3D detection, assuming input points $p \in \mathbb{R}^3$, F-PN uses a segmentation network to produce a segmentation mask determining background $0^{(p)}$ and car points $1^{(p)}$ and giving each a segmentation score $s^{(p)} \in (0,1)$. We minimize the maximum score given to a ``car" point $m = \max\limits_{p\in1^{(p)}} \{s^{(p)}\}$ using binary cross entropy, as shown in Eqn. (\ref{loss_fpn_pc}). Finally, following previous work \cite{uber}, we weighed the objective function by the intersection over union (IoU) score between the ground truth $y_{gt}$ and the predicted $\hat{y}$ bounding boxes: 
\vspace{-0.1cm}
\begin{equation}
\vspace{-0.1cm}
\mathcal{L}^{f-pn} = -1 \cdot log(1-m) \cdot IoU(y_{gt},\hat{y}) \label{loss_fpn_pc} 
\end{equation}

EPNet outputs many bounding box proposals for each detected vehicle; each proposal has a confidence score $s$. Similar to \cite{uber}, we aim to suppress relevant proposals $\hat{y} \in \mathcal{Y}$, where $\mathcal{Y}$ is the set of proposals with scores $s>0.1$. We use binary cross entropy again to minimize the scores given to proposals as shown in Eqn. \ref{loss_ep}. Similarly, we weigh the objective by the IoU score. This objective applies to both shape and texture optimization and runs end-to-end since this is a fusion model.
\vspace{-0.2cm}

\begin{equation}
\mathcal{L}^{epnet} = \frac{1}{|\mathcal{Y}|}\sum_{s, \hat{y} \in \mathcal{Y}} -1 \cdot log(1-s) \cdot IoU(y_{gt},\hat{y}) \label{loss_ep}. 
\end{equation}

\begin{figure}[htbp]
\centering
  \subfigure{
      \includegraphics[width=0.47\linewidth]{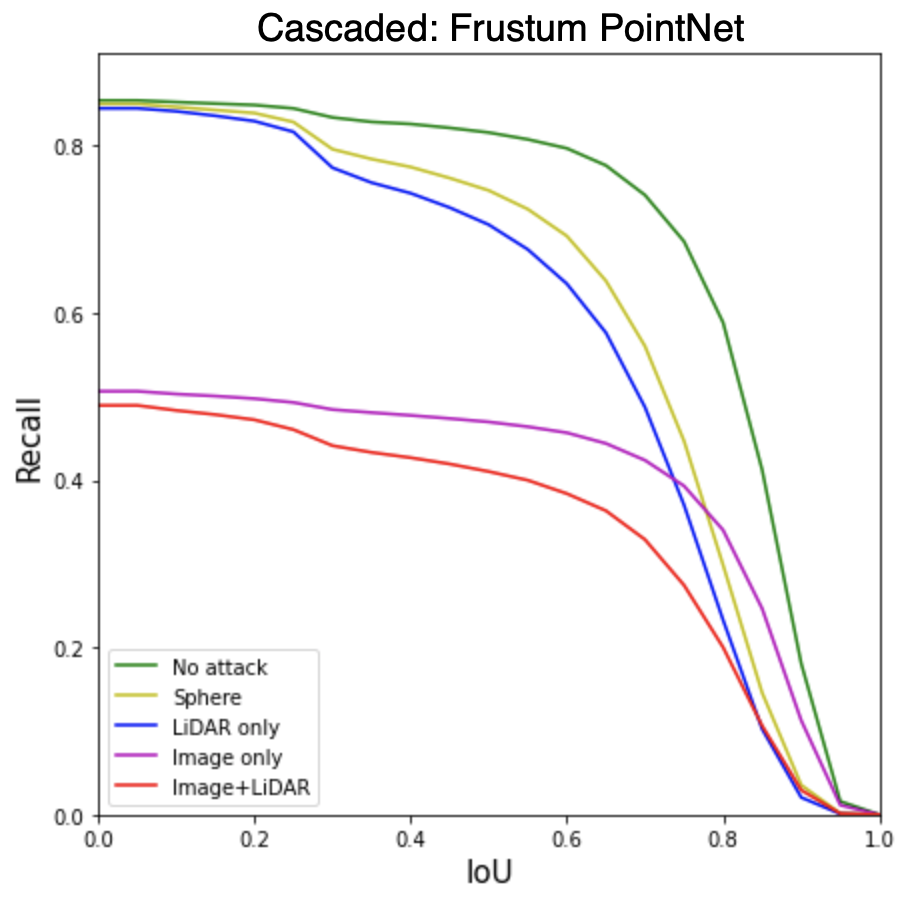}}
  \subfigure{
        \includegraphics[width=0.47\linewidth]{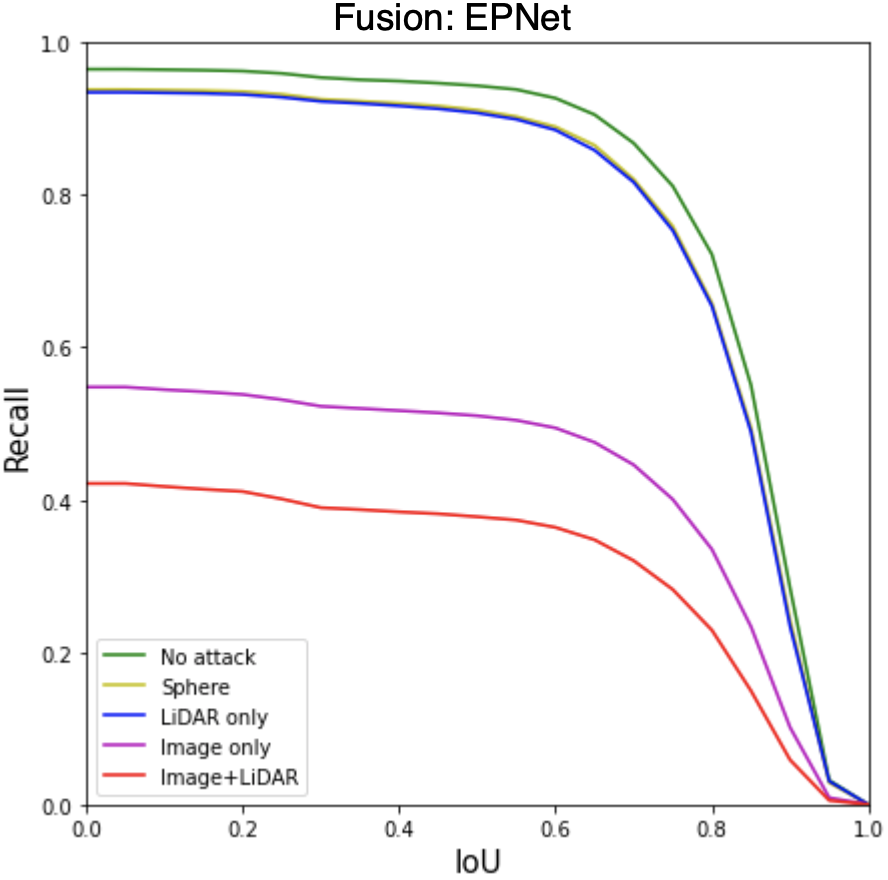}}
        \vspace{-0.3cm}
\caption{Victim vehicle recall results as the IoU threshold varies. We notice high recall in F-PN under low IoU thresholds, since it always assumes the existence of a car in a frustum. Also, the fusion model is much more robust to LiDAR-only attacks when compared to the cascaded model. However, both models are very vulnerable to image attacks.}
\label{bev_recall}
\vspace{-0.5cm}
\end{figure}

We also want to ensure that the geometry of the mesh is smooth and realistic, so we minimize a Laplacian loss \cite{liu2019softras}: $\mathcal{L}_{lap}=\sum_i\norm{\delta_i}^2_2 \label{lap}$, where $\delta_i$ is the distance between a vertex $v_i$ and the centroid of its neighbors $N(i)$.

\begin{equation}
    \delta_i = v_i -  \frac{1}{\norm{N(i)}}\sum_{j\in N(i)}v_j
\end{equation}
The overall adversarial attack objective function for EPNet is shown in Eqn. \eqref{loss_fus}, and for the point cloud pipeline in the cascaded F-PN in Eqn. \eqref{loss_casc}, where $\lambda$ is a weight coefficient for the Laplacian smoothing loss that is chosen empirically.

\vspace{-0.2cm}

\begin{equation}
\mathcal{L}^{epnet}_{adv} = \mathcal{L}^{epnet} + \lambda \mathcal{L}_{lap} \label{loss_fus}
\vspace{-0.2cm}
\end{equation}

\begin{equation}
\mathcal{L}^{f-pn}_{adv} = \mathcal{L}^{f-pn} + \lambda \mathcal{L}_{lap} \label{loss_casc}
\end{equation}

\vspace{-0.3cm}

\section{Experiments}
\label{sec:experiments}
\subsection{Experiment Setup}
We use the KITTI dataset \cite{kitti}, a popular benchmark for 3D detection in autonomous driving, to train the adversarial object. Similar to many prior work \cite{frustum, epnet} we use 3,712 LiDAR+image samples for training, and 3,769 for validation. All reported results in the following section are from the validation set. This is a universal attack, i.e., the shape and texture of the object are trained on the entire training set. 

The attack pipeline for the cascaded model is shown in Fig. \ref{pipe_frustum}. We divide our attack to 2 stages, as F-PN has two separate pipelines: RGB detection and point cloud detection. To attack F-PN's point cloud pipeline, we use an initial 40 cm radius isotropic sphere mesh with 162 vertices and 320 faces. Projected gradient descent is used to keep the object's size within an $80\times80\times80 \text{ }cm$ box. We use a small mesh because in cascaded models the point cloud pipeline takes fewer points (limited 3D space from 2D region proposal) than models that input the entire scene's point cloud. We use ground truth 2D bounding boxes to generate the frustums used in training and evaluating the attack on the point cloud pipeline. An ADAM optimizer is used to iteratively deform the mesh to minimize Eqn. \eqref{loss_casc}. Once we have an adversarial shape we move to the next stage: we render it to 2D images and use an ADAM optimizer to learn a universal adversarial texture for a multi-modal attack. As mentioned earlier, the texture is produced via interpolation of learned per-vertex colors. We use YOLOv3 generated 2D bounding boxes to measure the effect of an image-only attack and an image+point cloud attack on detection. In the cascaded case, the image-only or LiDAR-only attacks are mainly an ablation study since we only render to the modality under consideration. We use these attacks to see which input modality caused higher vulnerability.

\begin{figure}[t!]
 \center
  \includegraphics[scale=0.395]{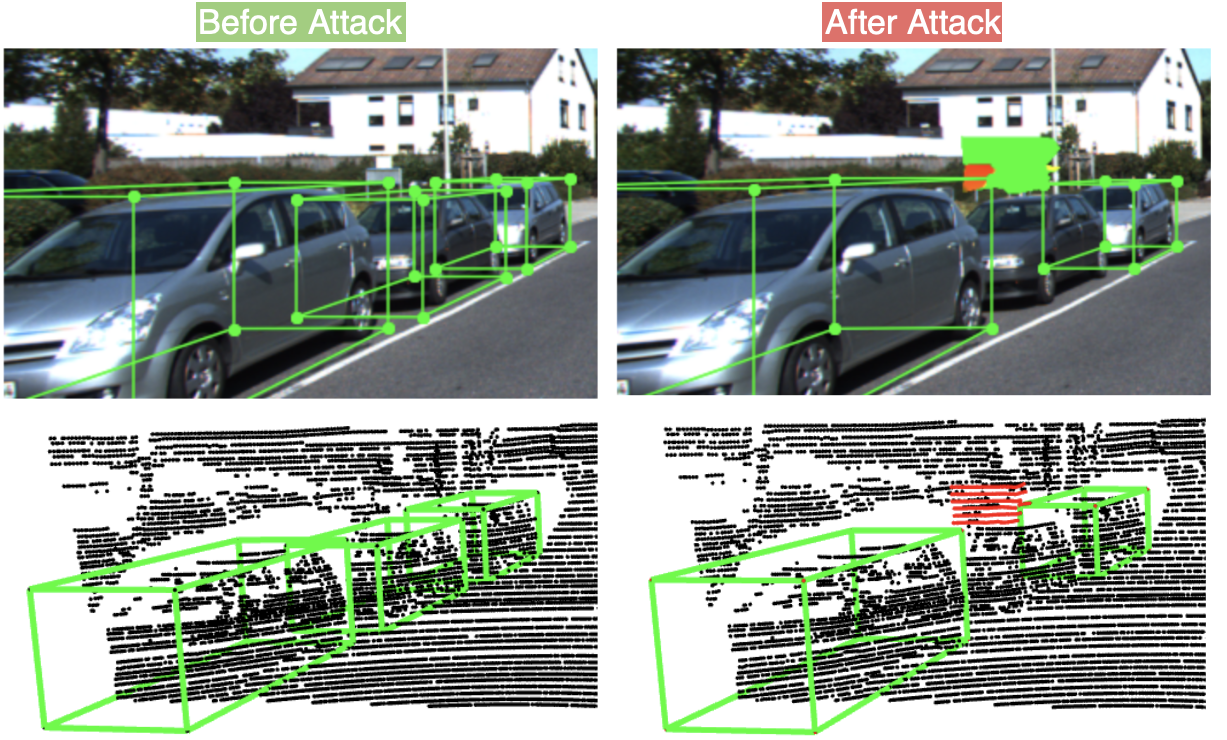}
  \caption{Example from the multi-modal attack on the fusion EPNet model \cite{epnet}. We show detection on the image and LiDAR point cloud before and after an attack. Note that cars with the adversarial mesh on them are undetected.}
  \label{examples}
  \vspace{-0.5cm}
\end{figure}

For the fusion model, we train the shape and texture simultaneously, and end-to-end as in Fig. \ref{pipe_fusion}. We use the same initial sphere mesh as in the F-PN attack, but since this model takes the whole 3D scene we allow a slightly larger adversarial mesh by keeping width and length under 120 cm and height under 80 cm. We also use projected gradient descent with an ADAM optimizer to learn the geometry and texture of the adversarial mesh. For the LiDAR-only attack, we learn an adversarial shape while keeping the texture a constant grey color. For the image-only attack, we use a constant sphere shape and learn the adversarial texture of this sphere. For the image+LiDAR attack, we allow adversarial perturbations to shape and texture.

In the results, we report the drop in a model's BEV \emph{average precision} (AP) after putting the mesh on all cars in the validation set. We show the results for 3 detection difficulty levels: easy, moderate, and hard (based on visibility and level of occlusion) with an IoU threshold of $0.7$. This can help compare our results with future attack and defense methods. Moreover, we can compare the model's AP on the KITTI benchmark before and after an attack. This can provide insight into the robustness of a perception model. We use the new KITTI evaluation protocol that uses 40 recall points instead of 11 to get clearer and fairer results.

We also report the \emph{attack success rate}, which is the percentage of detected vehicles that go undetected after placing the adversarial object. We count a vehicle detected if its IoU with the ground truth is greater than 0.7. We also report the recall of victim vehicles across different IoU thresholds. Recall is the number of cars detected over the number of all cars in the dataset. As a control we compare our adversarial attack results to just applying a grey spherical mesh instead of an adversarial mesh.

\begin{table}[htbp]
\begin{center}
\begin{tabularx}{\columnwidth}{ l|XXX }
\toprule
\textbf{Attack Type}&\textbf{Easy} & \textbf{Moderate} &\textbf{Hard} \\
\midrule
\textbf{\textit{No Attack (Clean)}} & 91.17 & 85.08 & 81.00\\
\textbf{\textit{No Attack (Sphere)}}& 52.81 & 51.46 & 48.19 \\
\textbf{\textit{LiDAR Only}} & 37.36 & 37.79 & 39.10  \\

\textbf{\textit{Image Only}} & 50.71 & 36.17 & 31.07 \\

\textbf{\textit{LiDAR + Image}} & \textbf{29.04} & \textbf{23.52} & \textbf{19.66} \\
\bottomrule
\end{tabularx}
\end{center}
\vspace{-0.1cm}
\caption{Cascaded: F-PN car detection AP results}
\label{fpn_pc_ap}
\vspace{-0.6cm}
\end{table}
\vspace{-0.3cm}

\begin{table}[htbp]
\begin{center}
\begin{tabularx}{\columnwidth}{ l|XXX }
\toprule
\textbf{Attack Type}&\textbf{Easy} & \textbf{Moderate} &\textbf{Hard} \\
\midrule
\textbf{\textit{No Attack (Clean)}} & 95.94 & 88.83 & 88.50\\
\textbf{\textit{No Attack (Sphere)}}& 94.74 & 87.47 & 84.58 \\
\textbf{\textit{LiDAR Only}} & 93.92 & 87.02 & 82.56  \\

\textbf{\textit{Image Only}} & 37.99 & 41.32 & 37.25 \\

\textbf{\textit{LiDAR + Image}} & \textbf{22.05} & \textbf{25.63} & \textbf{22.58} \\
\bottomrule
\end{tabularx}
\end{center}
\vspace{-0.1cm}
\caption{Fusion: EPNet car detection AP results}
\vspace{-1.0cm}
\label{epnet_ap}
\end{table}

\vspace{0.1cm}
\subsection{Results \& Discussion}

First, we find that a multi-modal adversarial attack lead to a drop in AP going from nearly 96\% to 22\%, and from 91\% to 29\% in the fusion and cascaded models respectively under the easy scenario as shown in Table \ref{fpn_pc_ap} and \ref{epnet_ap}. The multi-modal attack was successful in hiding more than half the cars from both models as shown in Table \ref{asr}. The sharp drop in AP can indicate that our attack introduced some false positives. Moreover, we mentioned in section \ref{sec:methods}, that EPNet can be thought of as the multi-modal fusion counterpart to the LiDAR-only PointRCNN model \cite{pointrcnn} which was attacked in \cite{uber} with an attack success rate of 32.3\%. While EPNet gained accuracy over PointRCNN because of incorporating image features using a novel fusion method, it became much more vulnerable to image-based adversarial attacks. As shown in Table \ref{asr}, a multi-modal adversarial attack on EPNet had an attack success rate of 63.19\%, a nearly 30\% gain over the attack on PointRCNN. An example of the attack on EPNet is shown in Fig. \ref{examples}.

We find that LiDAR-only attacks are much more effective on the cascaded model than the fusion one. From Fig. \ref{bev_recall}, we can see that, as the IoU thresholds increase, recall drops quickly with the insertion of any object (in this case a sphere) on top of a car even without an attack. Recall is reduced further after the insertion of our adversarial object. This is also reflected in the sharp decreases in AP under a 0.7 IoU threshold, as shown in Table \ref{fpn_pc_ap}. This can be attributed to the fact that cascaded models are mainly trained on frustum-shaped limited 3D spaces that just encapsulate a car. This means that it is not exposed to objects around or on cars. This poses a grave danger for autonomous driving, since correct estimation of dimensions and locations of surrounding vehicles and objects is important to autonomous decision making. On the other hand, fusion models are not much affected with a sphere or a LiDAR-only adversarial attack. This is probably because fusion models directly incorporate image features which can lend robustness against LiDAR-only adversarial attacks. Also, EPNet generates many proposals for a single car from foreground points and thus it can challenging to suppress them all in a LiDAR-only attack.

Both architectures are very vulnerable to adversarial attacks that target the image pipeline. In cascaded models, the 2D proposals decide exactly where to search for objects and so if that fails the entire model fails. Using image-based region proposals does not utilize a main purpose behind LiDAR which is to avoid cases where lighting and occlusion affect detection and localization. Also, RGB attacks are much simpler and less computationally expensive than point cloud attacks, and they are more easily reproduced in real life which can make them more dangerous. The fusion model as well was heavily affected because of the image features. As shown in Table \ref{asr}, an image-only attack on EPNet had a success rate of 48.64\%. This can be due to the known issue of brittle image features in DNNs. In the fusion case, adding a LiDAR attack to the image attack improved on the success rate by nearly 15\%. On the other hand, adding the LiDAR attack in the cascaded case improved the attack success rate by nearly 13\%. 

\begin{table}[t!]
\begin{center}
\begin{tabularx}{\columnwidth}{ l|XXX }
\toprule
\textbf{Attack Type}&\textbf{Cascaded: FP-N} & \textbf{Fusion: EPNet}\\
\midrule
\textbf{\textit{No Attack (Sphere)}}& 24.39\% & 5.45\% \\
\textbf{\textit{LiDAR Only}} & 34.24\% & 5.91\%  \\

\textbf{\textit{Image Only}} & 42.8\% & 48.64\% \\

\textbf{\textit{LiDAR + Image}} & \textbf{55.60\%} & \textbf{63.19\%} \\
\bottomrule
\end{tabularx}
\end{center}
\vspace{-0.1cm}
\caption{Attack success rate results for the two models}
\label{asr}
\vspace{-1.1cm}
\end{table}

\vspace{-0.15cm}
\section{Conclusion \& Future Work}
\vspace{-0.1cm}
We proposed a universal and physically realistic adversarial attack on multi-modal 3D detection models used in car perception. We manipulated mesh geometry and texture and used differentiable rendering to study the vulnerability of both representative cascaded and fusion camera-LiDAR models. We found that both model types are vulnerable to the proposed multi-modal adversarial attacks mainly due to the brittle image features. We also showed that the proposed attack can successfully make a car evade detection from the two studied models more than 50\% of the time.

Our attack methodology is generalizable and can be extended in several directions: For example, making the adversarial object more inconspicuous. We start from a 3D sphere, other works can start from other 3D shapes especially ones commonly found on cars like kayaks, or suitcases and keep the deformation limited using $\lambda$ from Eqn. \ref{loss_fus}. Some works can also use more complex texture representations and sampling from a real-world dataset to make the adversarial texture resemble naturally-appearing colors. Moreover, we train the adversarial object to produce false negatives, but it can also be trained to introduce any other error like false positives by tweaking the objective functions and using any other transformation matrix $\mathbf{T}$. Such an attack similarly poses great dangers to autonomous driving. Also, we attack two prominent models from two common architectures, but our methods (using a 3D mesh, rendering to image and point cloud, etc.) can be slightly modified to attack any other camera-LiDAR DNN or even conduct black-box adversarial attacks. Finally, this work can also be used to more thoroughly investigate adversarial defense methods and DNNs trained on out-of-distribution objects.
\vspace{-0.2cm}
\bibliographystyle{IEEEtran}
\bibliography{ref}

\end{document}